\documentclass[journal]{IEEEtran}

\ifCLASSINFOpdf
\else
\fi
\usepackage{times}
\usepackage{xcolor}
\usepackage{soul}
\usepackage[utf8]{inputenc}
\usepackage[small]{caption}
\usepackage{multirow}
\usepackage{graphicx}  
\usepackage{helvet}
\usepackage{courier}
\usepackage{url}
\usepackage{color}
\usepackage{amsmath}
\usepackage{amssymb}
\usepackage{algorithm}
\usepackage{algorithmic}

\usepackage{amsmath}

\def\ie{{\em i.e.}}
\def\eg{{\em e.g.}}

\begin{document}

%
\title{POSITIVE SEMIDEFINITE SUPPORT VECTOR REGRESSION METRIC LEARNING}
%

\author{ Lifeng Gu

\thanks{Tianjin University.}}

\markboth{IEEE Transactions on Neural Networks and Learning Systems}%
{Shell \MakeLowercase{\textit{et al.}}: Bare Demo of IEEEtran.cls for IEEE Journals}

\maketitle

\begin{abstract}
  Most existing metric learning methods focus on learning a similarity or distance measure relying on similar and dissimilar relations between sample pairs.
  However, pairs of samples cannot be simply identified as similar or dissimilar in many real-world applications, e.g., multi-label
  learning, label distribution learning.
  To this end, relation alignment metric learning (RAML)  framework is proposed to handle the metric learning problem in those scenarios. But RAML framework uses SVR solvers for optimization. It can't learn positive semidefinite distance metric which is necessary in metric learning. In this paper, we propose two methods to overcame the weakness. Further, We carry out several experiments on the single-label classification, multi-label classification, label distribution learning to demonstrate the new methods achieves favorable performance against RAML framework.
\end{abstract}


\IEEEpeerreviewmaketitle

\section{Introduction}
In many computer vision and pattern recognition tasks, \eg, face recognition \cite{guillaumin2009you}, image classification \cite{mensink2012metric}, and person re-identification \cite{liao2015person}, it is crucial to learn a discriminative distance metric to measure the similarity between pairs of samples.
Intuitively, metric learning aims to learn a discriminative similarity or dissimilarity metric by pushing the dissimilar samples away and pulling the similar samples together.
Typical distance metrics include Euclidean distance, cosine distance, and Mahalanobis distance.
Most existing metric learning methods focus on learning a discriminative Mahalanobis distance.
Beyond Mahalanobis distance, generalized distance metric learning methods are presented by learning high-order discriminant functions \cite{li2012beyond}.
According to the availability of the label information, metric learning can be partitioned into three categories, \ie, the unsupervised, semi-supervised and supervised methods.
To deal with the heterogeneous data, multi-modal \cite{mcfee2011learning} and cross-modal \cite{wang2016joint} metric learning algorithms are developed.
Because of the diversity of the feature space, linear, kernel and tensor distance metrics are learned for different data structures.
Different from shallow metric learning, deep learning based methods learn the feature and metric jointly and achieve superior performance \cite{oh2016deep}.

One of the key steps in existing metric learning methods is to generate doublet \cite{davis2007information}, triplet \cite{weinberger2009distance} or even quadruplet \cite{law2013quadruplet} constraints using the label information.
Doublet constraints are the most commonly used in metric learning methods.
Similar and dissimilar sample pairs are generated in the $k$-nearest neighbors or $\varepsilon$-neighborhood by measuring whether two samples belong to the same class.
In some applications, \eg, weakly supervised learning \cite{mu2010weakly} or social networks \cite{shaw2011learning}, sample pairs are generated from connectivity information or other side information.
Generally, there are two sets of sample pairs, \ie, one contains the similar sample pairs and the other one contains the dissimilar ones.

However, for some learning tasks, \eg, multi-label learning \cite{zhang2015lift} and label distribution learning \cite{geng2016label}, relations between sample pairs cannot be simply identified as similar or dissimilar. Thus, the existing metric learning methods cannot work on the above tasks.

The problem arises that it is difficult to classify two images into similar or dissimilar sample pair.
Above discussions encourage us to propose a generalized metric learning method, which can be flexibly adopted to various kinds of tasks.

In machine learning community, one of the basic assumptions is that samples should keep with the same relations in different spaces, especially in the feature space and label space.
The principle of metric learning is to encourage samples in the feature space to satisfy the expected relations induced by supervised information.
Manifold learning emphasizes locality preserving, which requires that the nearest neighbors of samples should be close to each other in the projected low-dimensional feature space \cite{yang2006efficient}.
For kernel learning machines, the kernel matrix can be considered as the similarity relation of all samples.
Kernel alignment exploits the similarity between kernel matrices for learning kernels \cite{cortes2012algorithms} and matrix completion \cite{bhadra2017multi}.
For multi-modal learning, the sample relation in feature spaces of different modalities should be consistent with that in the label space.
For metric learning, as long as the sample relations in the decision space are modeled, the distance metric can be learned by minimizing the difference between sample relations in feature space and decision space.

In this paper, we propose two metric learning formulation, namely RAML-PCSVR and RAML-NCSVR.
Our methods aims to overcome the limitations of RAML framework.
The contributions of this paper are summarized as follows.
\begin{itemize}
  \item Two metric learning formulation are proposed to learn distance metrics for different learning tasks, including single-label learning, multi-label learning, and label distribution learning.
  \item The proposed methods can learn positive semi-definite distance metric directly than RAML framework.
  \item Experiments on single-label classification, multi-label classification and label distribution learning tasks show that our RAML-PCSVR  and PAML-NCSVR achieves superior performance against RAML framework.
\end{itemize}

\section{Related Work}
For metric learning, doublet constraint is a kind of description of relationship between a pair of samples in the decision space.
 $f({\bf{x}}_i, {\bf{x}}_j, {\bf{M}},b)$ is used to measure the sample relations in feature space while $g({\bf{y}}_i,{\bf{y}}_j)$ is used to measure the sample relations in decision space.
$g({\bf{y}}_i,{\bf{y}}_j)$ is specially designed for different tasks.
Let $\bf{A} \in {\mathbb{R}}^{n \times n}$ and $\bf{E} \in {\mathbb{R}}^{n \times n}$ be the sample relation matrix in feature and decision spaces, respectively.
In general, sample relation in the feature space should be consistent with that in the decision space, \ie,
\begin{eqnarray}
\small
\left[ {\begin{array}{*{20}{c}}
{{a_{11}}}&{...}&{{a_{i1}}}&{...}&{{a_{n1}}}\\
{...}&{...}&{...}&{...}&{...}\\
{{a_{1i}}}&{...}&{{a_{ii}}}&{...}&{{a_{ni}}}\\
{...}&{...}&{...}&{...}&{...}\\
{{a_{1n}}}&{...}&{{a_{in}}}&{...}&{{a_{nn}}}
\end{array}} \right] = \left[ {\begin{array}{*{20}{c}}
{{e_{11}}}&{...}&{{e_{i1}}}&{...}&{{e_{n1}}}\\
{...}&{...}&{...}&{...}&{...}\\
{{e_{1i}}}&{...}&{{e_{ii}}}&{...}&{{e_{ni}}}\\
{...}&{...}&{...}&{...}&{...}\\
{{e_{1n}}}&{...}&{{e_{in}}}&{...}&{{e_{nn}}}
\end{array}} \right]
\nonumber
\end{eqnarray}
where $a_{ij}$ and $e_{ij}$ represent the sample relation of ${\bf{x}}_i$ and ${\bf{x}}_j$ in the feature space and decision space, respectively.
Here, to keep consistency, we require that
\begin{equation}
    f({\bf{x}}_i, {\bf{x}}_j, {\bf{M}},b)=g({\bf{y}}_i,{\bf{y}}_j).
\label{consistency}
\end{equation}
where $g({\bf{y}}_i,{\bf{y}}_j)$ is the difference degree of two samples in the decision space.
$g({\bf{y}}_i,{\bf{y}}_j)$  reflects the sample relation in the decision space, and guides the learning of $({\bf{M}},b)$ in feature space.

\begin{equation}
\begin{array}{l}
f({{\bf{x}}_i},{{\bf{x}}_j},{\bf{M}},b) = {\left( {{{\bf{x}}_i} - {{\bf{x}}_j}} \right)^T}{\bf{M}}\left( {{{\bf{x}}_i} - {{\bf{x}}_j}} \right) + b\\
{\rm{                      }}  \qquad  \qquad  \qquad  \; = \left\langle {{\bf{M}},{{\bf{T}}_{ij}}} \right\rangle  + b
\end{array}
\end{equation}
where $\left\langle {\cdot ,\cdot } \right\rangle $ is defined as the Frobenius inner product of two matrices, $b$ is the bias item, and ${{\bf{T}}_{ij}} = \left( {{{\bf{x}}_i} - {{\bf{x}}_j}} \right){\left( {{{\bf{x}}_i} - {{\bf{x}}_j}} \right)^T}$.
Then we rewrite \eqref{consistency} to
\begin{equation}
    g({{\bf{y}}_i},{{\bf{y}}_j}) = \left\langle {{\bf{M}},{{\bf{T}}_{ij}}} \right\rangle  + b
    \label{linearregression}
\end{equation}
Once the relation function $g({{\bf{y}}_i},{{\bf{y}}_j})$ is chosen, \eqref{linearregression} can be considered as a linear regression problem.
Hence, the metric learning problem is converted to solve a sample pair regression problem with the scaled second sample moment ${\bf{T}}_{ij}$ of sample pair $({\bf{x}}_i,{\bf{x}}_j)$ as the input.

\subsection{Sample Pair Kernel}
To formulate the sample pair regression problem in \eqref{linearregression}, \cite{zhu2018} introduce a $2$-degree polynomial kernel for sample pairs.
Let ${\bf{z}}_i$ denote the sample pair $({{\bf{x}}_{i1}},{{\bf{x}}_{i2}})$.
Then the $2$-degree polynomial kernel is defined as
\begin{equation}
\begin{array}{l}
k({{\bf{z}}_i},{{\bf{z}}_j}) = \left\langle {{{\bf{T}}_i},{{\bf{T}}_j}} \right\rangle \\
 = tr\left( {\left( {{{\bf{x}}_{i1}} - {{\bf{x}}_{i2}}} \right){{\left( {{{\bf{x}}_{i1}} - {{\bf{x}}_{i2}}} \right)}^T}\left( {{{\bf{x}}_{j1}} - {{\bf{x}}_{j2}}} \right){{\left( {{{\bf{x}}_{j1}} - {{\bf{x}}_{j2}}} \right)}^T}} \right)\\
 = {\left( {{{\left( {{{\bf{x}}_{i1}} - {{\bf{x}}_{i2}}} \right)}^T}\left( {{{\bf{x}}_{j1}} - {{\bf{x}}_{j2}}} \right)} \right)^2}
\end{array}
\end{equation}
With the sample pair kernel, given a sample pair ${\bf{z}}=({{\bf{x}}}_1,{{\bf{x}}}_2)$, the regression function can be rewritten as
\begin{equation}
   f({\bf{z}}) = \sum\nolimits_{i = 1}^n {{\beta_i}\left\langle {{\bf{T}},{{\bf{T}}_i}} \right\rangle }  + b=\left\langle  {\bf{M}}, {\bf{T}}  \right\rangle+b
\end{equation}
where ${\bf{T}} = \left( {{{\bf{x}}_1} - {{\bf{x}}_2}} \right){\left( {{{\bf{x}}_1} - {{\bf{x}}_2}} \right)^T}$ and ${{\bf{T}}_i} = \left( {{{\bf{x}}_{i1}} - {{\bf{x}}_{i2}}} \right){\left( {{{\bf{x}}_{i1}} - {{\bf{x}}_{i2}}} \right)^T}$.
Here ${\bf{M}} = \sum\nolimits_{i= 1}^n {{\beta _i}{{\bf{T}}_i}}$.
$\bf{M}$ is actually a linear combination of the scaled sample moments of the difference between two samples in one pair.

\section{Support Vector Regression Metric Learning}
In this section, we first review RAML framework, then we will point out it's limitations.
\cite{zhu2018} propose RAML formulation  to develop a SVR-like distance metric method:
\begin{equation}
\begin{array}{l}
\mathop {\min }\limits_{\bf{M},{\xi},\xi^*} {\rm{ }}  \lambda r({\bf{M}}) +  {\rho ({\xi},\xi^*)} \\
s.t.\left\{ \begin{array}{l}
g({\bf{z}}_i) - \left( {\left\langle {{\bf{M}},{{\bf{T}}_i}} \right\rangle  + b} \right) \le \varepsilon  + {\xi _i}\\
 \left( {\left\langle {{\bf{M}},{{\bf{T}}_i}} \right\rangle  + b} \right)-g({\bf{z}}_i)  \le \varepsilon  + \xi _i^*\\
\xi _i^*,{\xi _i} \ge 0
\end{array} \right.
\end{array}
\label{krl}
\end{equation}
where ${\xi _i}$ and $\xi _i^*$ are slack variables,
and ${\rho ({\xi},\xi^*)}$ is the margin loss item.
By using Frobenius norm regularization for $r({\bf{M}})$ and $\varepsilon$-sensitive loss function for $\rho ({\xi _{}},\xi _{}^*)$, the metric learning problem in \eqref{krl} can be formulated as:
\begin{equation}
\begin{array}{l}
\mathop {\min }\limits_{\bf{M},{\xi},\xi^*} \frac{{\rm{1}}}{{\rm{2}}} \left\|\bf{M}\right\|_F^2 + \lambda \sum\nolimits_{i = 1}^n ({{\xi _i} + \xi _i^*}) \\
s.t.\left\{ \begin{array}{l}
g({{\bf{z}}_i}) - \left( {\left\langle {{\bf{M}},{{\bf{T}}_i}} \right\rangle  + b} \right) \le \varepsilon  + {\xi _i}\\
\left( {\left\langle {{\bf{M}},{{\bf{T}}_i}} \right\rangle  + b} \right)-g({{\bf{z}}_i}) \le \varepsilon  + \xi _i^*\\
\xi _i^*,{\xi _i} \ge 0
\end{array} \right.
\end{array}
\label{mlsvr}
\end{equation}
where $\left\| {\bf{M}} \right\|_F^2$ is the Frobenius norm  of $\bf{M}$, and $\lambda$ is a trade-off constant.
By using the Lagrange multipliers, we have
\begin{equation}
{\rm{L = }}\left\{ \begin{array}{l}
\frac{{\rm{1}}}{{\rm{2}}}\left\| {\bf{M}} \right\|_F^2 + \lambda \sum\nolimits_{i = 1}^n ({{\xi _i} + \xi _i^*})  - \\
\sum\nolimits_{i = 1}^n {{a_i}\left( {\varepsilon  + {\xi _i} - g({{\bf{z}}_i}) + \left\langle {{\bf{M}},{{\bf{T}}_i}} \right\rangle  + b} \right)}  - \\
\sum\nolimits_{i = 1}^n {a_i^*\left( {\varepsilon  + \xi _i^* + g({{\bf{z}}_i}) - \left\langle {{\bf{M}},{{\bf{T}}_i}} \right\rangle  - b} \right)}  - \\
\sum\nolimits_{i = 1}^n {\left( {{\eta _i}{\xi _i} + \eta _i^*\xi _i^*} \right)}
\end{array} \right\}
\label{svrdualorgi}
\end{equation}
All dual variables should satisfy the positivity constraints, \ie, ${a_i},a_i^*,{\eta _i},\eta _i^* \ge 0$.
According to the saddle point condition, the partial derivatives of $L$ with respect to the primal variables will be vanishing, \ie,
\begin{equation}
    \frac{{\partial {\rm{L}}}}{{\partial b}}{\rm{  =   }}\sum\nolimits_{i = 1}^n {\left( {{a_i} - a_i^*} \right)}  = 0
    \label{de1}
\end{equation}
\begin{equation}
   \frac{{\partial {\rm{L}}}}{{\partial {\bf{M}}}}{\rm{  =  }}{\bf{M}} - {\rm{ }}\sum\nolimits_{i = 1}^n {\left( {{a_i} - a_i^*} \right)} {{\bf{T}}_i} = 0
    \label{de2}
\end{equation}
\begin{equation}
   \frac{{\partial {\rm{L}}}}{{\partial \xi _i^*}}{\rm{  =  \lambda}} - a_i^* - \eta _i^*
    \label{de3}
\end{equation}
Substituting \eqref{de1}, \eqref{de2} and \eqref{de3} into \eqref{svrdualorgi}, we get the dual optimization problem of \eqref{mlsvr} with
\begin{equation}
\begin{array}{l}
\max \left\{ \begin{array}{l}
{\rm{ - }}\frac{1}{2}\sum\limits_{i,j = 1} {\left( {{a_i} - a_i^*} \right)\left( {{a_j} - a_j^*} \right)} \left\langle {{{\bf{T}}_i},{{\bf{T}}_j}} \right\rangle \\
 - \varepsilon \sum\limits_{i = 1}^n {\left( {{a_i} + a_i^*} \right)}  + \sum\limits_{i = 1}^n {g({{\bf{z}}_i})\left( {{a_i} - a_i^*} \right)}
\end{array} \right\}\\
s.t.\sum\limits_{i = 1}^n {g({{\bf{z}}_i})\left( {{a_i} - a_i^*} \right)}  = 0,{a_i},a_i^* \in \left[ {0,\lambda} \right]
\end{array}
\label{svrdual}
\end{equation}

Similar to the solution of SVR, we can get the solution for \eqref{svrdual}, \ie,
\begin{equation}
{\bf{M}} = \sum\nolimits_{i = 1}^n {\left( {{a_i} - a_i^*} \right){{\bf{T}}_i}}
\label{M_eq}
\end{equation}

Then, the corresponding regression function can be rewritten as
\begin{equation}
 f({\bf{z}}) = \sum\nolimits_{i = 1}^n {\left( {{a_i} - a_i^*} \right)} \left\langle {{{\bf{T}}_i},{{\bf{T}}}} \right\rangle  + b
\end{equation}
For the metric learning task, $\bf{M}$ is required to be positive semi-definite.
Whereas, the solution for \eqref{svrdual} cannot ensure that $\bf{M}$ is a PSD matrix.
\cite{zhu2018} compute the singular value decomposition of $\bf{M}={\bf{U\Lambda V}}$ and only keep the positive part of $\bf{\Lambda}$ to form a new matrix ${\bf{\Lambda}}_{+}$.
Finally, the PSD matrix $\bf{M}= \bf{U} {\bf{\Lambda}}_{+} \bf{V}$.


\section{Positive Semidefinite Support Vector Regression Metric Learning}
In order to use efficient SVR solvers instead of quadratic programming solvers to speed up algorithm, the $\bf{M}$ learned by \eqref{svrdual} is not a PSD matrix,\cite{zhu2018}  transform it simplified by singular value decomposition to get a PSD matrix, but it will heart the discriminating power of $\bf{M}$. Now we propose{ {RAML-PCSVR} }and {{RAML-NCSVR}}, we describe how to learn a PSD matrix directly by SVR.\\
RAML-PCSVR is easy: In \eqref{M_eq}, if we make sure ${a_i} \geq a_i^*,i = 1,2,\dots,n, $ the $\bf{M}$ learned will be a PSD matrix. We can easily prove it.\\

   Denote by $\bf{\mu} \in {\mathbb{R}^{d}}$ a random vector, We have:  \\
   \begin{equation}
   \begin{array}{l}
     {\bf{\mu}}^{T}\bf{M\mu} = {\bf{\mu}}^{T}\left(\sum\nolimits_{i = 1}^n {\left( {{a_i} - a_i^*} \right){{\bf{T}}_i}}\right){\bf{\mu}}\\\\
     =  \bf{\mu}^{T}\left(\sum\nolimits_{i = 1}^n \left( {{a_i} - a_i^*} \right){\left( {{{\bf{x}}_{i1}} - {{\bf{x}}_{i2}}} \right){\left( {{{\bf{x}}_{i1}} - {{\bf{x}}_{i2}}} \right)^T}}\right)\bf{\mu}\\\\
     = \sum\nolimits_{i = 1}^n{\left({{a_i} - a_i^*}\right)}{\bf{\mu}}^{T}\left({{{\bf{x}}_{i1}} - {{\bf{x}}_{i2}}}
     \right){\bf{\mu}}\left({{{{\bf{x}}_{i1}} - {{\bf{x}}_{i2}}}}^{T}\right)\\\\
     = \sum\nolimits_{i = 1}^n{\left({{a_i} - a_i^*}\right)\left({\bf{\mu}}^{T}\left({{{\bf{x}}_{i1}} - {{\bf{x}}_{i2}}}\right)\right)^{2}}
     \end{array}
     \label{proof_eq}
       \end{equation}

Since $\left({{a_i} - a_i^*}\right)\left({{{\bf{x}}_{i1}} - {{\bf{x}}_{i2}}}\right)^{2} \geq 0$,  ${\bf{\mu}}^{T}{\bf{M}}\bf{\mu}\geq 0$,  therefore $\bf{M}$ is a PSD matrix. Our optimization objective becomes 
\begin{equation}
\begin{array}{l}
\max \left\{ \begin{array}{l}
{\rm{ - }}\frac{1}{2}\sum\limits_{i,j = 1} {\left( {{a_i} - a_i^*} \right)\left( {{a_j} - a_j^*} \right)} \left\langle {{{\bf{T}}_i},{{\bf{T}}_j}} \right\rangle \\
 - \varepsilon \sum\limits_{i = 1}^n {\left( {{a_i} + a_i^*} \right)}  + \sum\limits_{i = 1}^n {g({{\bf{z}}_i})\left( {{a_i} - a_i^*} 
 \right)}
\end{array} \right\}\\
s.t.\sum\limits_{i = 1}^n {g({{\bf{z}}_i})\left( {{a_i} - a_i^*} \right)}  = 0,{a_i},a_i^* \in \left[ {0,\lambda} \right],
{a_i} \geq a_i^*
\end{array}
\label{psd_svrdual}
\end{equation}
\eqref{psd_svrdual} just modified \eqref{svrdual} by add constraints, it can be solved by quadratic programming. It's slower than best SVR solvers.\\
Now, we introduce  {{RAML-NCSVR}}, trying a different way to learn a PSD matrix through RAML  formulation instead of modifying it's dual problem, we define $\bf{M} = \sum\nolimits_{i=1}^{n}{\mu_{i}}{\bf{T}}_{i}$, where $\mu_{i}$ is the scalar combination coefficient and $\mu_{i} \geq  0$, similar than \eqref{proof_eq}, $\bf{M}$ is a PSD matrix, by substituting ${\bf{M}}$ with   
$\sum\nolimits_{i=1}^{n}{\mu_{i}}{\bf{T}_{i}}$, we write the  new  formulation:
\begin{equation}
\begin{array}{l}
\mathop {\min }\limits_{{\mu},{\xi},{\xi^*}} \frac{{\rm{1}}}{{\rm{2}}} {\sum\limits_{i,j = 1}\mu_{i}\mu_{j}\left\langle{\bf{T_{i}}},{\bf{T_{j}}}\right\rangle}  + \lambda \sum\nolimits_{i = 1}^n ({{\xi _i} + \xi _i^*}) \\
s.t.\left\{ \begin{array}{l}
g({{\bf{z}}_i}) - {\sum\limits_{j=1}\mu_{j}{\left\langle {{{\bf{T}}_j},{{\bf{T}}_i}} \right\rangle}  }  \le \varepsilon  + {\xi _i}\\
\sum\nolimits_{j=1}\mu_{j}{\left\langle {{{\bf{T}}_j},{{\bf{T}}_i}} \right\rangle} -g({{\bf{z}}_i}) \le \varepsilon  + \xi _i^*\\
\xi _i^*,{\xi _i} \ge 0, \mu_{i} \ge 0
\end{array} \right.
\end{array}
\label{psd_svr_prime}
\end{equation}
By introducing the Lagrange multipliers, it's Lagrangian is:
\begin{equation}
 {\rm{ L = }} \left\{
\begin{array}{l}
  
    \frac{{\rm{1}}}{{\rm{2}}} {\sum\limits_{i,j = 1}\mu_{i}\mu_{j}\left\langle{\bf{T_{i}}},{\bf{T_{j}}}\right\rangle} 
    + \lambda \sum\nolimits_{i = 1}^n ({{\xi _i}+ \xi _i^*})\\
    +\sum\nolimits_{i=1}\alpha_{i}\left(g({{\bf{z}}_i}) - {\sum\nolimits_{j=1}\mu_{j}{\left\langle {{{\bf{T}}_j},{{\bf{T}}_i}} \right\rangle}  }   -\varepsilon  - {\xi _i}\right)\\
    +\sum\nolimits_{i=1}\alpha^{*}_{i}\left(\sum\nolimits_{j=1}\mu_{j}{\left\langle {{{\bf{T}}_j},{{\bf{T}}_i}} \right\rangle}  -g({{\bf{z}}_i}) - \varepsilon  - \xi _i^*\right)\\
    -\sum_{i=1}{\eta_{i}\xi_{i}}
    -\sum_{i=1}{\eta_{i}^{*}\xi_{i}^{*}}
    -\sum_{i=1}{\sigma_{i}\mu_{i}}
\end{array}\right\}
\label{psd_lagrange}
\end{equation}
where $\alpha_{i}, \alpha_{i}^{*}, \eta_{i}, \eta_{i}^{*}, \sigma_{i}$ are the Lagrange multipliers, which satisfied$\alpha_{i} \ge 0, \alpha_{i}^{*}\ge 0, \eta_{i}\ge 0, \eta_{i}^{*}\ge 0, \sigma_{i} \ge 0.$ The paritial derivatives of L with respect to the primal variables are:
\begin{equation}
\begin{array}{l}
    \frac{\partial{L}}{\partial u_{i}}=\sum_{j=1}{u_{j}-(\alpha_{j}-\alpha{j}^{*})}-{\sum\left(\alpha_{i}-\alpha_{i}^{*}\right)\left\langle\bf{T_{i},T_{j}}\right\rangle} \\- {\sigma_{i}} =0
     
\end{array} 
\label{dldu}
\end{equation}
\begin{equation}
\begin{array}{l}
    \frac{\partial{L}}{\partial   \xi_{i}}=\lambda-\alpha_{i}-\eta_{i}=0
\end{array}    
\end{equation}
\begin{equation}
\begin{array}{l}
    \frac{\partial{L}}{\partial   \xi_{i}^{*}}=\lambda-\alpha^{*}-\eta_{i}^{*}=0
\end{array}    
\end{equation}
In order to solve \eqref{dldu}, we introduce a auxiliary variable $\rho$, which satisfies $\sigma_{i} $ = $\sum_{j=1}\rho_{j}\left\langle\bf{T_{i}},\bf{T_{j}}\right\rangle$, \eqref{dldu} becomes
\begin{equation}
\begin{array}{l}
    \sum_{j=1}\left({\mu_{j}-\left(\alpha_{j}-\alpha_{j}^{*}\right)-\rho_{j}}\right)\left\langle\bf{T_{i}},\bf{T_{j}}\right\rangle = 0\\
\end{array} 
\label{dldu1}
\end{equation}
Because$ \left\langle\bf{T_{i}},\bf{T_{j}}\right\rangle \ge 0$, so we have:

\begin{equation}
   \mu_{j}-\left(\alpha_{j}-\alpha_{j}^{*}\right)-\rho_{j} = 0  
\end{equation}
\begin{equation}
    \mu_{j} = \left(\alpha_{j}-\alpha_{j}^{*}\right)+\rho_{j}
\label{mu_update}    
\end{equation}

Substituting above back into \eqref{psd_lagrange},  we get
the following Lagrange dual problem:
\begin{equation}
\begin{array}{l}

\max\limits_{\rho,\alpha,\alpha^{*}} \left\{
\begin{array}{l}
\frac{{\rm{1}}}{{\rm{2}}} \sum\limits_{i,j = 1}{\left(\alpha_{i}-\alpha_{i}^{*}+\rho_{i}\right)}{{\left(\alpha_{j}-\alpha_{j}^{*}+\rho_{j}\right)}\left\langle{\bf{T_{i}}},{\bf{T_{j}}}\right\rangle} \\
    + \lambda \sum\nolimits_{i = 1}^n ({{\xi _i}+ \xi _i^*})
    +\sum\nolimits_{i=1}\alpha_{i}(g\left({{\bf{z}}_i}\right) -\\ 
    {\sum\nolimits_{j=1}{\left(\alpha_{j}-\alpha_{j}^{*}+\rho_{j}\right)}{\left\langle {{{\bf{T}}_j},{{\bf{T}}_i}} \right\rangle}}   -\varepsilon  - {\xi _i})\\
        +\sum\nolimits_{i=1}\alpha^{*}_{i}(\sum\nolimits_{j=1}{\left(\alpha_{j}-\alpha_{j}^{*}+\rho_{j}\right)}{\left\langle {{{\bf{T}}_j},{{\bf{T}}_i}} \right\rangle} \\
    -g({{\bf{z}}_i}) - \varepsilon  - \xi _i^*)
    -\sum_{i=1}{\eta_{i}\xi_{i}}\\
    -\sum_{i=1}{\eta_{i}^{*}\xi_{i}^{*}}
    -\sum_{i=1}{\sigma_{i}{\left(\alpha_{j}-\alpha_{j}^{*}+\rho_{j}\right)}}

  \end{array}\right\}\\
s.t. {a_i},a_i^* \in \left[ {0,\lambda} \right], \sum_{j=1}\rho_{j}\left\langle\bf{T_{i}},\bf{T_{j}}\right\rangle \ge 0
\end{array}
\label{psd_svrdual1}
\end{equation}
There are three groups variables in \eqref{psd_svrdual1}, we adopt an alternative optimization approach to
solve them. They can be solved by quadratic programming. First, given $\rho$, the variables $\alpha$ and $\alpha^{*}$ can be solved 
as follows:    
\begin{equation}
    \begin{array}{l}
         \max\limits_{\alpha,\alpha^{*}}\left\{
         \begin{array}{l}
        -\frac{1}{2}\sum_{i=1}\sum_{j=1}\left(\alpha_{i}-\alpha_{i}^{*}\right)\left(\alpha_{j}-\alpha_{j}^{*}\right)\left\langle{\bf{T_{i}},\bf{T_{j}}}\right\rangle\\
        +\sum\nolimits_{i=1}\left(\alpha_{i}-\alpha_{i}^{*}\right)g({{\bf{z}}_i})-\sum_{i=1}(\alpha_{i}-\alpha_{i}^{*})
        \\
        \sum_{j=1}\rho_{j}\left\langle{\bf{T_{i}},\bf{T_{j}}}\right\rangle
          \end{array}\right\}\\
s.t.  {a_i},a_i^* \in \left[ {0,\lambda} \right]
    \end{array}
    \label{alpha_update}
\end{equation}

Then, given the variables $\bf{\alpha}$ and $\bf{\alpha^{*}}$, $\bf{\rho}$ can be solved 
as follows:    
\begin{equation}
    \begin{array}{l}
         \max\limits_{\rho}\left\{
         \begin{array}{l}
        -\frac{1}{2}\sum_{i=1}\sum_{j=1}\rho_{i}\rho_{j}\left\langle{\bf{T_{i}},\bf{T_{j}}}\right\rangle\\
        -\sum_{i=1}\sum_{j=1}(\alpha_{i}-\alpha_{i}^{*})
        \rho_{j}\left\langle{\bf{T_{i}},\bf{T_{j}}}\right\rangle
        \end{array}\right\}\\
s.t. \sum_{j=1}\rho_{j}\left\langle\bf{T_{i}},\bf{T_{j}}\right\rangle \ge 0
    \end{array}
    \label{rho_update}
\end{equation}
We summarize them in Algorithm \ref{alg:psd_raml}
\begin{algorithm}[t]
\caption{The algorithms of our proposed RAML-PCSVR and PRML-NCSVR}
\begin{algorithmic}[1]
\REQUIRE ~~\\
{Training data $\bf{X} \in {\mathbb{R}}^{d \times m}$, where $d$ and $m$ are the numbers of feature dimension and samples, respectively.\\
       }
\STATE Generate sample pairs $({\bf{x}}_{i1},{\bf{x}}_{i2})$,$i=1,2,...,n$.\\
\STATE Compute sample relation $g({\bf{x}}_{i1},{\bf{x}}_{i2})$,$i=1,2,...,n$.\\
\STATE RAML-PCSVR: Solve \eqref{psd_svrdual1} by quadratic programming\\
       RAML-NCSVR:
\REPEAT{
   \STATE Update $\bf{\alpha}$ and $\bf{\alpha^{*}}$ by \eqref{alpha_update}  \\
        
   \STATE Update $\bf{\rho}$  by \eqref{rho_update}
   \STATE Update $\bf{\mu}$  by \eqref{mu_update}
       }
\UNTIL{converge}
\STATE RAML-PCSVR: ${\bf{M}} = \sum\nolimits_{i = 1}^n {\left( {{a_i} - a_i^*} \right){{\bf{T}}_i}}$.\\
       RAML-NCSVR: ${\bf{M}} = \sum\nolimits_{i = 1}^n {\mu_{i}{{\bf{T}}_i}}$.\\     
\ENSURE ~~\\

Distance metric matrix ${\bf{M}}$\\
\end{algorithmic}
\label{alg:psd_raml}
\end{algorithm}

\section{Discussions}

\subsection{Sample Relation Function}
We reuse the sample relation function in RAML. The motivation of RAML is keeping relation consistency in different spaces, including feature space and label space.
As the sample relations in the decision space are used to guide the metric learning in feature space, it is important to choose proper sample relation functions for different kinds of decision spaces. We consider four learning tasks, \ie, single label learning, multi-label learning, label distribution learning and regression task.
Let ${\bf{y}}_i$ and ${\bf{y}}_j$ denote the label vector of ${\bf{x}}_i$ and ${\bf{x}}_j$.
The sample relation function is defined as:
\begin{equation}
    g({{\bf{y}}_i},{{\bf{y}}_j}){\rm{  = }}{\left\| {{{\bf{y}}_i} - {{\bf{y}}_j}} \right\|_1}
\label{relationg}
\end{equation}
where ${\left\| {\bf{a}} \right\|_1}{\rm{ }}$ is the $l_1$-norm of $\bf{a}$.
For single label classification, when $g({{\bf{y}}_i},{{\bf{y}}_j})$ is defined as \eqref{relationg}, RAML degenerates to a sample pair classification problem.
For multi-label learning, \eqref{relationg} reflects the difference with respect to positive classes of two samples.
For label distribution learning, there are many metrics to evaluate the difference between two distributions. For        regression, it reflects the difference between two continues value.
Here, we experimentally find that \eqref{relationg} reflects sample difference in the decision space and achieves superior performance.
Therefore, we choose \eqref{relationg} in RAML-SVR, RAML-KRR  for all learning tasks. 
The choice of optimal relation functions for different tasks are still an open problem, which will be investigated in our future work.
If we want to learn a similarity metric in feature space, the inner product of two vectors, or other
kernel functions can be used for $g({{\bf{y}}_i},{{\bf{y}}_j})$.

\subsection{Sample Pair Selection}
Sample pair selection method is not changed in our methods. Relation alignment learning aims to preserve the consistency of the sample relations between the feature space and the decision space.
However, we do not need to use the relations of all sample pairs.
\begin{table*}[htbp!]
\centering
\tabcolsep=0.03in
\begin{tabular}{c|cccccc}

\hline
  &RAML-SVR &RAML-PCSVR &RAML-NCSVR\\
\hline
distance matrix \bf{M} &$\sum\nolimits_{i = 1}^n {\left( {{a_i} -a_i^*}\right){{\bf{T}}_i}}$
& $ \sum\nolimits_{i = 1}^n {\left( {{a_i} - a_i^*} \right){{\bf{T}}_i}}$
&$\sum\nolimits_{i = 1}^n {\left( {{a_i} - a_i^*}+\rho_{i} \right){{\bf{T}}_i}}$

\\
\hline

\multirow{2}{*}{constraints}
&${a_i}\geq 0, a_i^*\geq 0$
&${a_i}\geq 0, a_i^*\geq 0 $
&${a_i}\geq 0, a_i^*\geq 0, \rho_{i} \in {\mathbb{R}}  $
\\&$ {a_i} -a_i^*\geq 0$
&${{a_i} - a_i^*}+\rho_{i} \geq 0$
\\
\hline
loss function&
$\varepsilon$-sensitive &$\varepsilon$-sensitive &$\varepsilon$-sensitive 

\\
\hline
regularization & Frobenius norm& Frobenius norm& Frobenius norm&
\\
\hline
\end{tabular}

\caption{connections and differences between different RAML formulations}
\label{connectionstable}
\end{table*}

\begin{table*}[htbp!]
\centering
 \tabcolsep=0.03in
 \resizebox{\textwidth}{!}{
    \begin{tabular}{ccccccccccc}
   \hline
    Data &S/F/C  & ITML  & LDML  & LMNN  & DSVM & GMML  & DML   & RAML-SVR & RAML-PCSVR & RAML-NCSVR \\\hline
    binalpha&1404/320/36  & 0.6303$ \pm $0.0501 & 0.6542$ \pm $0.0317 & 0.6112$ \pm $0.0358 & 0.5625$ \pm $0.0322 & 0.5338$ \pm $0.1986 & 0.5063$ \pm $0.0251 & \textbf{{0.7250$ \pm $0.0348}} &  \textbf{0.7296$ \pm $0.0386}& \textbf{\underline{0.7315$\pm$0.0349} } \\
    caltech101&8641/256/101  & \textbf{0.5803$ \pm $0.0162} & 0.5528$ \pm $0.0157 & 0.5795$ \pm $0.0126 & 0.5584$ \pm $0.0159 & 0.5500$ \pm $0.0117 & 0.3936$ \pm $0.0123 & \textbf{{0.5855$ \pm $0.0095}} &\textbf{0.6065$\pm$0.1824}&\textbf{\underline{0.6128$\pm$0.0174}}\\
    MnistDat&3495/784/10  & 0.8695$ \pm $0.0142 & 0.8858$ \pm $0.0124 & 0.8721$ \pm $0.0255 & 0.8848$ \pm $0.0194 & 0.8589$ \pm $0.0171 & 0.8323$ \pm $0.0239 & \textbf{0.9019$ \pm $0.0175} 
    & \textbf{{0.9272$\pm$0.0109}}& \textbf{\underline{0.9330$\pm$0.0987}}\\
    Mpeg7 &1400/6000/70  & 0.8214$ \pm $0.0333 & 0.7971$ \pm $0.0365 & 0.8253$ \pm $0.0232 & 0.8271$ \pm $0.0353 &\textbf{ 0.8429$ \pm $0.0228 }& 0.7071$ \pm $0.0267 & \textbf{{0.8450$ \pm $0.0305}} & 
    \textbf{{0.8529$\pm$0.02569}}&\textbf{\underline{0.8536$\pm$0.02802}}\\
    news20&3970/8014/4  & 0.8678$ \pm $0.0200 & 0.8816$ \pm $0.0145 & 0.8734$ \pm $0.0290 & 0.8594$ \pm $0.0159 & 0.8647$ \pm $0.0143 & 0.8166$ \pm $0.0222 & \textbf{0.9025$ \pm $0.0132} & \textbf{{0.9245$ \pm $0.0165}}&\textbf{{0.8864$ \pm $0.0752}}\\
    TDT2\_20&1938/3677/20  & 0.9587$ \pm $0.0358 & 0.9531$ \pm $0.0306 & 0.9352$ \pm $0.0197 & 0.9499$ \pm $0.0175 & 0.9437$ \pm $0.0275 & 0.6333$ \pm $0.0176 & \textbf{0.9679$ \pm $0.0244} & 
     \textbf{0.9845$\pm$0.0198}&\textbf{\underline{0.9875$\pm$0.0101}}\\
    uspst&2007/256/10  & 0.8979$ \pm $0.0261 & 0.9084$ \pm $0.0243 & 0.9096$ \pm $0.0217 & 0.9125$ \pm $0.0172 & 0.8858$ \pm $0.0168 & 0.8030$ \pm $0.0330 & \textbf{{0.9525$ \pm $0.0147}} & \textbf{0.9477$\pm$0.0177}&  \textbf{{0.9519$\pm$0.0165}}\\
  \hline
 \end{tabular}
 }
 \vspace{-0.01cm}

  \caption{Classification accuracy on image datasets }
   \label{table2}
\end{table*}%

For support vector regression, the support vectors are mainly lying on the decision boundary.
Therefore, sample pairs are only generated in the $k$ nearest neighbors , which is similar to most existing metric learning algorithms.
Besides, using only part of sample pairs can greatly reduce computational complexity and storage burden.
\section{Connections Between Different RAML Formulations}
We find there are connections between RAML and our proposed methods. when we get the solution of $\bf{M}$.  \eqref{psd_svrdual} modified \eqref{svrdualorgi} by add constrains in Lagrange multipliers
$\bf{\alpha}$ and $\bf{\alpha^{*}}$ so that we can get a PSD matrix directly, auxiliary variables $\rho$ introduced in \eqref{psd_svr_prime} relax the constraints in \eqref{psd_svrdual} and we can get a better matrix, we summary their connections in \ref{connectionstable} \\

\section{Experiments}
In this section, we conduct experiments to validate the performance of the proposed metric learning methods.
We consider three applications, including single-label classification, multi-label classification, label distribution learning
. The following part will be organized as the corresponding parts.
\subsection{Single-Label Classification}
\noindent\textbf{Experiment setup.}
The detailed information of  datasets is listed in Table  Table \ref{table2}, where "S/F/C" represents the number of samples, features and classes.
We compare our methods with the state-of-the-art methods, \ie, ITML \cite{davis2007information}, LMNN \cite{weinberger2009distance}, DML \cite{ying2012distance}, DoubletSVM (DSVM) \cite{wang2015kernel},,GMML \cite{zadeh2016geometric} on each dataset.
For fair comparison, the parameters of all compared methods are set as the default setting of the original
references.
For DSVM, we set $k = 1$, and the penalty factor $C <10,000$.
For GMML,  the weight $t$ is set within [0,1]  and chosen by greedy search.
Ten-fold cross validation is introduced to evaluate the metric learning performance, \ie, 90\% for training and 10\% for testing.
The average accuracy of 10-fold cross validation is reported.

\noindent\textbf{Experimental analysis.}
 Table \ref{table2} list the classification accuracy of  different metric learning methods on image datasets, respectively, where the best results are marked in bold face.
RAML-SVR  indicate support vector regression metric learning, 
Our methods achieves superior results in terms of the evaluation criteria on most dataset. RAML-PCSVR and RAML-NCSVR both perform better than RAML-SVR in all dataset, approximation operation in RAML-SVR heart the discrimination power of distance matrix, but RAML-SVR is much faster than RAML-PCSVR and RAML-NCSVR. It can be used to process big dataset.
For RAML-KRR, when the number of samples increase significantly, the efficiency will be reduced because its time complexity is $o(n^3)$, where $n$ is the number of samples.

%
%

\subsection{Multi-Label Classification}
\textbf{Datasets.}
In this section, we evaluate the proposed method using three datasets \footnote{http://mulan.sourceforge.net/datasets-mlc.html}, \ie, emotion \cite{trohidis2008multi}, flags, and corel800 dataset \cite{hoi2006learning}.
The emotion dataset \cite{trohidis2008multi} consists of 100 songs from each of the following 7 different genres, \emph{Classical, Reggae, Rock, Pop, Hip-Hop, Techno and Jazz}.
The collection was created from 233 musical albums choosing three songs from each album.
The flag dataset contains 194 instances, 19 features and 7 labels (red, green, blue, yellow, white, black, orange).
The corel 800 dataset \cite{hoi2006learning} contains 800 grayscale images of 10 individuals with 80 images per class.
\\
\noindent\textbf{Evaluation metrics.}
We employ five multi-label classification measures as evaluation metrics including Hamming loss, ranking loss, one error, coverage and average precision.
Hamming loss measures accuracy in a multi-label classification task.
Ranking loss has the property that the minimization of the loss functions will lead to the maximization of the ranking measures.
MLKNN is the multi-label version of KNN \cite{zhang2007ml} and it is based on statistical information derived from the label sets of an unseen instance’s neighboring instances.
As no specific metric learning algorithms are developed for MLKNN, here we use MLKNN as the baseline.
If the performance of RAML is superior to MLKNN, the effectiveness of RAML is verified.
\noindent\textbf{Experimental analysis.}
Experimental results of RAML and  MLKNN are reported in Table \ref{mlknnresult}, where the best result on each evaluation criterion is shown in bold face.
The "$\downarrow$" after the measures indicates ``the smaller the better`` and "$\uparrow$" after the measures indicates ``the larger the better``.
As shown in Table \ref{mlknnresult},  both RAML-SVR, RAML-PCSVR and RAML-NCSVR achieve superior results in terms of the five evaluation measures.
Compared with MLKNN, RAML can learn a discriminative distance metric, making the sample relation in the feature space more consistent with that in the decision space. RMAL-NCSVR perform best.

\subsection{Label Distribution Learning}
\textbf{Datasets.}
The dataset employed in this experiment includes 2,000 natural scene images \cite{zhang2007ml}.
There are nine possible labels associated with these images, \ie, \emph{plant, sky, cloud, snow, building, desert, mountain, water and sun}.
The image features are extracted using the method in \cite{boutell2004learning}.
Each image is represented by a feature vector of 294 dimensions.
The output of each instance is a distribution rather than discrete labels.
AAKNN is the extended version of KNN in label distribution learning.
Here AAKNN is used as the baseline without metric learning in the label distribution task.

\noindent\textbf{Evaluation metrics.}
Different from both the single label output and the label set output of multi-label learning, the output of label distribution learning algorithm is a label distribution.
The evaluation measures for label distribution learning is the average distance or similarity between the predicted and real label distributions.
On a particular dataset, each of the measures may reflect some aspects of an algorithm.
It is hard to say which evaluation metric is the best.
Therefore, we use several measures to evaluate the proposed algorithm, and compare RAML and our methods with the classical AAKNN method.
Finally we employ five measures: Chebyshev distance (Cheb), Clark distance (Clark), Canberra metric (Canber), cosine coefficient (Cosine), and intersection similarity(Intersec) \cite{cha2007comprehensive}.
The first three are distance measures and the last two are similarity measures.

\noindent\textbf{Experimental analysis.}
Table \ref{ldlrsult} shows RAML and AAKNN in terms of five measures.
We show the best result with respect to each measure in bold face.
The "$\downarrow$" after the measures indicates ``the smaller the better``. "$\uparrow$"  after the measures indicates ``the larger the better``.
RAML-PCSVR and RAML-NCSVR perform better than AAKNN in terms of five different measures.
It owes to more discriminative metric learned by the proposed methods.
\begin{table}[H]\small
\centering
\tabcolsep=0.01in
\begin{tabular}{c|c|cccc}
\hline
&{ Data }& emotion & flags  & corel800 \\
\hline
\multirow{5}{*}{ MLKNN} &{Hamming Loss}$\downarrow$& 0.2137  & 0.3099    &  0.0137  \\
                        &{Ranking Loss}$\downarrow$& 0.1729  & 0.2228   &  0.1888  \\
                        &{One Error}$\downarrow$   & 0.3317  & 0.2154    &  0.6825  \\
                        &{Coverage}$\downarrow$    & 1.9158  & 3.8154    & 88.5100  \\
                        &{Average Precision}$\uparrow$ & 0.7808  & 0.8084    &  0.3276  \\

\hline
\multirow{5}{*}{ RAML-PCSVR}   &{Hamming Loss}$\downarrow$& \textbf{0.2103}  & \textbf{{0.2901}}    &  \textbf{0.0135}  \\
                        &{Ranking Loss}$\downarrow$& \textbf{0.1551} & \textbf{{0.2053}}    &  {0.1893}  \\
                        &{One Error}$\downarrow$   & \textbf{{0.2722}}  & \textbf{\underline{0.1692}}    &  \textbf{{0.6675}}  \\
                        &{Coverage}$\downarrow$    & \textbf{1.8317}  &\textbf{ 3.7692}    & \textbf{{88.5400}}  \\
                        &{Average Precision}$\uparrow$  & \textbf{0.8052}  & \textbf{{0.8244}}    &  \textbf{0.3297}  \\
                        \hline
\multirow{5}{*}{ RAML-SVR}   &{Hamming Loss}$\downarrow$& \textbf{0.2054}  & \textbf{{0.2967}}    &  \textbf{0.0135}  \\
                        &{Ranking Loss}$\downarrow$& \textbf{0.1577} & \textbf{0.2179}    &  \textbf{\underline{0.1882}}  \\
                        &{One Error}$\downarrow$   & \textbf{\underline{0.2376}}  & \textbf{{0.2000}}    &  \textbf{\underline{0.6425}}  \\
                        &{Coverage}$\downarrow$    & \textbf{1.8960}  &\textbf{ 3.8115}    & \textbf{{88.2350}}  \\
                        &{Average Precision}$\uparrow$  & \textbf{0.8101}  & \textbf{{0.8128}}    &  \textbf{0.3386}  \\
                                                \hline

\multirow{5}{*}{ RAML-NCSVR}   &{Hamming Loss}$\downarrow$& \textbf{\underline{0.1955}}  & \textbf{\underline{0.2549}}    &  \textbf{0.0135}  \\
                        &{Ranking Loss}$\downarrow$& \textbf{0.1560} & \textbf{\underline{0.1967}}    & {{0.1891}}  \\
                        &{One Error}$\downarrow$   & \textbf{{0.2723}}  & \textbf{{0.2000}}    &  \textbf{{0.6700}}  \\
                        &{Coverage}$\downarrow$    & \textbf{1.8168}  &\textbf{\underline {3.6769}}    & \textbf{\underline{88.1775}}  \\
                        &{Average Precision}$\uparrow$  & \textbf{0.8044}  & \textbf{\underline{0.8283}}    &  \textbf{0.3294}  \\

\hline
\end{tabular}
\caption{The performance of RAML-SVR, RAML-PCSVR, RAML-NCSVR, HRAML and MLKNN in terms of five evaluation measures.}
\vspace{-0.3cm}
\label{mlknnresult}
\end{table}.
\begin{table}[H]\footnotesize
\centering
\tabcolsep=0.01in

\begin{tabular}{c|ccccccc}
\hline
Criterion &{Chebyshev}$\downarrow$                            & {Clark}$\downarrow$         &{Canberra}$\downarrow$    &{Cosine}$\uparrow$  & {Intersection}$\uparrow$    \\
\hline
  AAKNN   & 0.3261  & 1.8448 & 4.3412 & 0.6905 & 0.5506\\
 RAML-PCSVR&{\textbf{0.3097}}  & \textbf{1.8160} & \textbf{4.2435} & \textbf{0.7077}  & \textbf{0.5739}\\
 RAML-NCSVR&{\textbf{0.3092}}  & \textbf{1.8186} & \textbf{4.2509} & \textbf{0.7050}  & \textbf{0.5676}\\
 RAML-SVR& {\textbf{0.3102}}  & \textbf{1.6986} & \textbf{3.8576} & \textbf{0.7051}  & \textbf{0.5739}\\

\hline
\end{tabular}
\caption{The performance of RAML-SVR, RAML-PCSVR, RAML-NCSVR and AAKNN in terms of five measures on Nature Scene dataset.}
\label{ldlrsult}
\end{table}
\vspace{-0.3cm}


\section{Conclusions}
In this paper, we proposed two methods to learn distance metrics for various kinds of learning tasks.
Different from RAML, our methods can learn positive semidefinite distance metric directly. 
Experimental result show RAML-PCSVR and RAML-NCSVR are very competitive with state-of-the-art metric learning methods on single-label classification, moreover they can improve the performance of multi-label learning, label distribution learning, and they perform better than RAML in most datasets.


\section{Acknowledgments}
This work was supported by the Tianjin university.

\bibliographystyle{IEEEtran}
\bibliography{egbib}
\ifCLASSOPTIONcaptionsoff
  \newpage
\fi

\end{document}